\pdfoutput=1

\documentclass[11pt]{article}

\usepackage[final]{acl}

\usepackage{times}
\usepackage{latexsym}

\usepackage[T1]{fontenc}

\usepackage[utf8]{inputenc}

\usepackage{microtype}

\usepackage{inconsolata}

\usepackage{graphicx}

\title{Developing a Reliable, Fast, General-Purpose Hallucination Detection and Mitigation Service}

\author{
 \textbf{Song Wang},
 \textbf{Xun Wang},
 \textbf{Jie Mei},
 \textbf{Yujia Xie},
\\
 \textbf{Sean Muarray},
 \textbf{Zhang Li},
 \textbf{Lingfeng Wu},
 \textbf{Si-Qing Chen},
 \textbf{Wayne Xiong}
\\
Microsoft
\\
 \{sonwang, xunwang, jimei, yujiaxie,\}@microsoft.com\\
 \{murraysean, z67li, lingfw, sqchen, weixi\}@microsoft.com
}

\begin{document}
\maketitle
\begin{abstract}

Hallucination, a phenomenon where large language models (LLMs) produce output that is factually incorrect or unrelated to the input, is a major challenge for LLM applications that require accuracy and dependability. In this paper, we introduce a reliable and high-speed production system aimed at detecting and rectifying the hallucination issue within LLMs. Our system encompasses named entity recognition (NER), natural language inference (NLI), span-based detection (SBD), and an intricate decision tree-based process to reliably detect a wide range of hallucinations in LLM responses. Furthermore, we have crafted a rewriting mechanism that maintains an optimal mix of precision, response time, and cost-effectiveness. We detail the core elements of our framework and underscore the paramount challenges tied to response time, availability, and performance metrics, which are crucial for real-world deployment of these technologies. Our extensive evaluation, utilizing offline data and live production traffic, confirms the efficacy of our proposed framework and service.

\end{abstract}

\section{Introduction}

In the rapidly evolving landscape of natural language processing (NLP), large language models (LLMs) have marked a significant leap forward, unlocking new horizons of capabilities and potentials. However, alongside their remarkable advancements, LLMs bring forth substantial challenges, with "hallucination" standing out as a particularly problematic issue. Hallucination in this context refers to instances when an LLM produces output that is either factually incorrect or not anchored in the supplied input, thus compromising the model's reliability and the credibility of its applications. Therefore, the importance of confronting and mitigating hallucinations in LLM deployments cannot be overstated.


Detecting and mitigating hallucinations present tough challenges, actively explored in recent research as evidenced by several survey papers \citep{Ji_2023, huang2023surveyhallucinationlargelanguage, Tonmoy2024ACS}. There are also different levels of hallucinations spanning from minor inconsistencies to blatant fabrications, and they can have different effects for different applications and users. Against this backdrop, our work delves into the nuances of hallucinations within LLMs, placing a special emphasis on \textbf{intrinsic hallucinations} i.e. errors that can be checked against reference inputs.

\begin{figure*}[t]
    \centering
\includegraphics[width=\linewidth]
{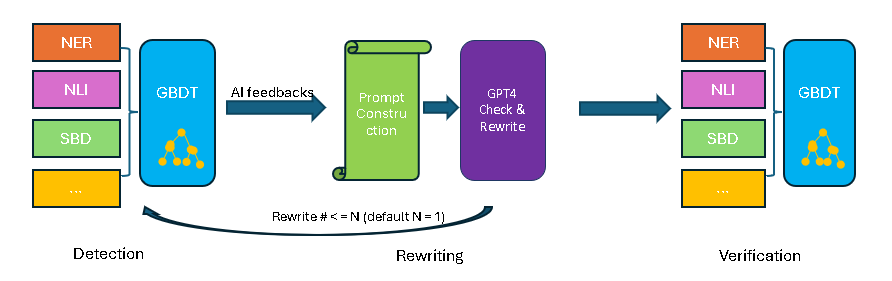}
    \caption{End-2-end hallucination detection and mitigation system}
    \label{fig:federated_pipeline}
\end{figure*}

Developing a \textbf{general-purpose, fast and accurate} hallucination detection and mitigation service is an extremely difficult task given the existing state-of-the-art technologies. To this end, we present a pragmatic solution as shown in Figure \ref{fig:federated_pipeline}, which includes three modules: \textbf{multi-source detection}, \textbf{iterative rewriting} and \textbf{multi-source verification}. We will discuss the components in details in Section \ref{sec:detection-mitigation-pipeline}.


Our contributions are as follows. First, we present a novel detection system capable of detecting different types of hallucinations with high accuracy. The system operates in real-time (low latency) and is suitable for large-scale applications (low cost). This approach leverages multiple hallucination detection methods—including named entity recognition (NER), natural language inference (NLI), and span-based detection—and ensembles multiple AI feedbacks using Gradient Boosting Decision Trees (GBDT).

Second, we propose a rewriting system for hallucination removal utilizing large language models (LLMs). After testing various strategies, we developed an effective rewriting solution that balances quality and latency.

Third, we conducted comprehensive experiments, analyses, and evaluations, demonstrating that our methods are effective and providing insights valuable for other researchers and industry scientists. The results are convincing and highlight the applicability of our methods in real-world scenarios.

\section{Related Work}
\paragraph{Hallucination Taxonomy}
A widely-adopted classification of hallucination is the intrinsic-extrinsic dichotomy\cite{dziri2021neural,huang2021factual}. Intrinsic hallucination occurs when LLM outputs contradict the provided input, such as prompts. Conversely, extrinsic hallucination occurs when LLM outputs cannot be verified by the information in the input. Recently, reasearchers have proposed more fine-grained taxonomies \cite{pagnoni-etal-2021-understanding,mishra2024finegrainedhallucinationdetectionediting}. 

We largely followed the categories in \cite{pagnoni-etal-2021-understanding} with modifications that reflect the nature and causes of hallucinations in LLM outputs and we also developed a guidelines based on this taxonomy for annotators. 

\paragraph{Hallucination Detection} 
Conventional methods of detecting hallucination can be classified into two types: token-based and sentence-based. The former aims to find hallucinated tokens while the latter is to identify the sentences with hallucinations. Various methods have been developed for identifying hallucinations and most of them leverage the pre-trained LLMs fine tuned on task-specific data \cite{liu2021token,dziri2021neural,cao2021hallucinated,zha2023alignscore}. More recently,
LLM with prompt-based methods are also widely used \citep{manakul2023selfcheckgptzeroresourceblackboxhallucination,lei2023chainnaturallanguageinference}. Though the prompt-based methods require little to no tuning and have competitive performance, they tend to have higher cost and higher latency.  

In our study, we've harnessed the strengths of both methods by employing LLM-based detection for labeling data and creating an ensemble of tailored traditional models such as NER, NLI, and span-based sequence labeling. Our designed detection service conducts a detailed examination of the input text’s semantic and syntactic attributes, allowing it to detect various kinds of hallucinations across different granularities and categories.

\paragraph{Hallucination Mitigation}
Hallucination mitigation is to correct the identified hallucinations in the generated
responses by LLMs. 
There are many other ways to reduce hallucination during post-generation. 
For example RARR \cite{thorne2021evidencebasedfactualerrorcorrection} trained a T5 model using retrieved evidence to generate corrected responses. More recently, researchers have been leveraging LLMs to better utilize hallucination feedback and generate corrections \cite{mündler2024selfcontradictoryhallucinationslargelanguage,dhuliawala2023chainofverificationreduceshallucinationlarge, lei2023chainnaturallanguageinference}.

In our work, our rewriter is also LLM-based, leveraging LLM's self refinement through feedback and reasoning. The key difference is that we have to take into consideration the cost and latency, which demands fewer output tokens, while ensuring the mitigation performance.

\begin{table*}[!ht] 
    \centering
\includegraphics[width=\linewidth]{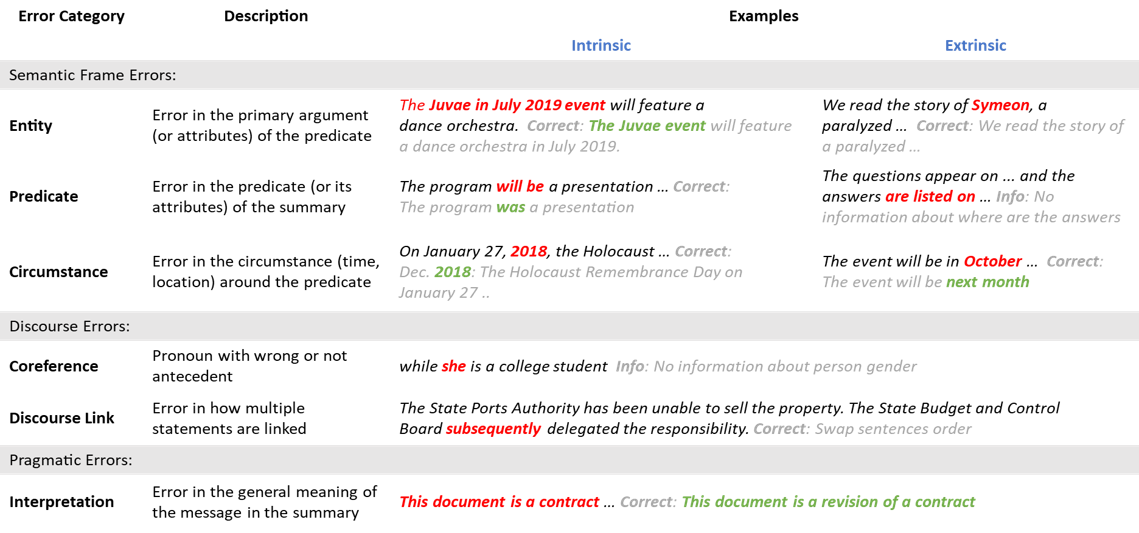}
    \caption{Hallucination taxonomy and examples. We largely followed the categories in \cite{pagnoni-etal-2021-understanding} with modifications that reflect the nature and causes of hallucinations in LLM outputs.}
    \label{tab:taxonomy}
\end{table*}

\begin{table}[ht!]
    \centering
\includegraphics[width=0.9\linewidth]{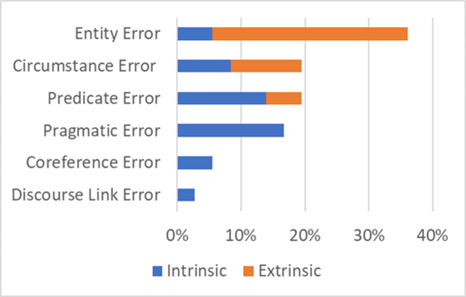}
    \caption{Hallucination distribution}
    \label{tab:taxonomy_dist}
\end{table}

\section{Hallucination Taxonomy and GPT4-based detection}

\subsection{Hallucination Taxonomy}
We started with developing a hallucination taxonomy by manually analyzing various hallucinated model outputs, mainly from some internal summarization systems, which is based on a state-of-the-art encoder-decoder model about 1B parameters. 

We randomly collected 500 samples from production systems and benchmark systems, including ChatGPT, and manually identified the \textbf{34} hallucinated outputs (two of the authors). In Table \ref{tab:taxonomy}, we list our taxonomy and examples of hallucinations, which show that hallucination is of different types and root causes. Most of them are due to semantic frame errors, but discourse errors and pragmatic errors also exist. 

In Table \ref{tab:taxonomy_dist}, we list the distribution of different types of hallucinations. As stated above, we focus mainly on \textbf{ intrinsic hallucinations}, which are errors that can be verified from the source document. Although most of the existing hallucination detection solutions are entity-based which treat new entities in the generated summary as hallucinations, from table \ref{tab:taxonomy_dist}, we found only a small portion, ~5\% of intrinsic hallucinations are attributable to new entities. This observation motivates us to develop an ensemble-based solution that extends NER-based methods to recall more hallucination errors.


\subsection{GPT4-based Detection}
\label{sec:gpt4-detector}
We developed a GPT4-based hallucination detection as follows (we cannot share the prompt due to proprietary limitations):  First, we transform the LLM outputs so that individual sentences are placed on separate lines. We then instruct GPT-4 to evaluate each sentence by comparing it against the source document and provide reasons if a sentence is determined to be a hallucination. If any sentence within the output is identified as hallucinated, the entire output is considered hallucinated. This approach utilizes the Chain of Thought (CoT) technique, and has been shown to outperform existing methods on multiple datasets; for instance, see the results on SummAC in Table \ref{tab:comparing_methods}. Other studies \cite{lei2023chainnaturallanguageinference, wei2024longformfactualitylargelanguage} have also demonstrated the effectiveness of leveraging GPT-4’s reasoning capabilities for hallucination detection.

Additionally, we examined discrepancies between human annotation and our GPT-4-based evaluation using 20 model outputs from a benchmark dataset of 1,400 samples (see Table \ref{tab:gpt4-vs-human}). Our analysis indicates that while GPT-4 tends to have higher false positive rates, human annotators often show higher false negative rates. Nonetheless, GPT-4's labeling is comparable to human efforts in overall error counts and can enhance annotator productivity by combining the model's high recall with human precision. However, the study's limited sample size and the specific design of prompts may constrain the generalizability of these findings, indicating a need for further research.

\begin{table}[ht]
  \centering
  \small
  \begin{tabular}{l|cc|c}
    \hline
     & \textbf{False Positive} & \textbf{False Negative} & \textbf{Total} \\
    \hline
    Annotators& 1 & 10 &11 \\
    GPT4& 4 & 5 & 9 \\
   \hline
  \end{tabular}
  \caption{Error analysis of the inconsistency of GPT4 and human,}
  \label{tab:gpt4-vs-human}
  \end{table}


\section{Hallucination Detection and Mitigation}
\label{sec:detection-mitigation-pipeline}
In this section, we introduce the hallucination checking method as shown in Figure \ref{fig:federated_pipeline} tailored for intrinsic hallucinations, which employes an ensemble method that leverages multiple techniques—including named entity recognition (NER), natural language inference (NLI), and sequence labeling—to detect inaccuracies. None of these models are LLM-based mainly for two main reasons: First, we aim to achieve real-time detection with high accuracy suitable for large-scale applications. The substantial costs associated with calling LLM models are not feasible. Second, LLMs are not effective at detecting hallucinations in their own outputs. Previous works have shown that LLMs tend to believe their outputs are correct and are difficult to persuade otherwise \cite{farquhar2024detecting,quevedo2024detecting}.

\subsection{Hallucination detection methods}
\begin{figure}[t]
    \centering
\includegraphics[width=0.5\linewidth]{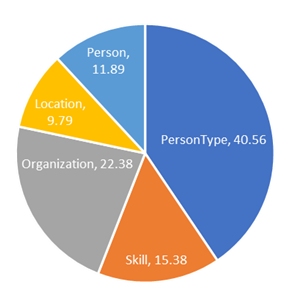} 
    \caption{Hallucination distribution over entity types}
    \label{fig:ner_entities}
\end{figure}

\paragraph{NER-based detection}
Named Entity Recognition (NER) aims to identify and categorize key information (entities) in text. By applying NER analysis to the input data, we can spot possible entities that are present in the LLM outputs but not supported by the source document—that is, hallucinations. We are using a well-known NER service\footnote{Azure AI Service: https://azure.microsoft.com/en-us/products/ai-services/ai-language} which returns both entity types and their confidence scores. We apply this NER service to detect hallucinations, and Figure \ref{fig:ner_entities} shows the common entity types among the hallucinated LLM outputs. Additionally, we conducted NER analysis on benchmark datasets in the target domain to determine the entity types and confidence thresholds for our NER detection implementation.

\paragraph{NLI-based detection }
In natural language inference (NLI), given two input text snippets—a premise and a hypothesis—the task is to predict their relationship: entailment, contradiction, or neutral. In principle, this aligns with the goal of hallucination detection. However, in most existing NLI datasets \citep{bowman-etal-2015-large, N18-1101, nie2020adversarialnlinewbenchmark, schuster-etal-2021-get}, the premises and hypotheses are short (one or two sentences). To address this, we included a new document-sentence dataset \citep{kamoi2023wicerealworldentailmentclaims} and used GPT-4, as described in Section \ref{sec:gpt4-detector}, to label a diverse set of document-summary pairs from various public and internal sources. Finally, we fine-tuned the pre-trained DeBERTa encoder \cite{he2021debertadecodingenhancedbertdisentangled} on these combined datasets. This model detects hallucinations based on the semantic relationship between the document and summary.

\paragraph{Span-based detection}
NLI provides sentence- or summary-level hallucination detection, while NER is restricted to a predefined set of named entities. To explore a more general fine-grained hallucination detection, we train a token-level hallucination detection model to provide more detailed AI feedback, such as highlighting hallucinated text spans.

\begin{figure}[!ht]
    \centering
    \includegraphics[width=\linewidth]{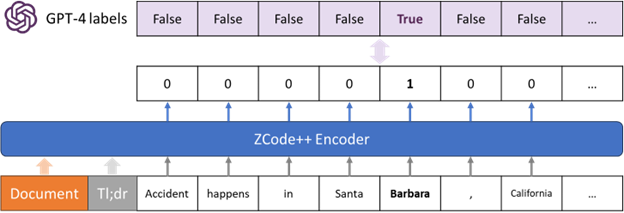}
    \caption{Span-based hallucination detection}
    \label{fig:rtd-intro}
\end{figure}

Starting with the dataset labeled by GPT-4 for the NLI model, as described in Section \ref{sec:gpt4-detector}, we further ask GPT-4 to highlight the hallucinated text spans if the text contains hallucinations. As shown in Figure \ref{fig:rtd-intro}, we initiate model training with a pre-trained Replaced Token Detection (RTD) head from DeBERTa \cite{he2021debertadecodingenhancedbertdisentangled}. We adapt this model using the GPT-4 generated data to determine if a token is part of a hallucinated span. We refer to this model as the Span-Based Detection model (SBD).

\subsection{Multi-Source Ensemble}

We believe that combining multiple sources is essential to further boost performance while minimizing costs. This approach is also convenient when we want to use a single score to control thresholds, allowing us to prioritize high precision during real implementation.

For this part, we collected about 10,000 training examples, where the labels (hallucinated vs. not hallucinated) are produced by GPT-4 detection, and the features are entities and their confidence scores from the NER, confidence score from NLI, and the confidence score from SBD. We adopt the \textbf{Gradient Boosting Decision Tree} to leverage the diverse AI feedback and fine-tune a model using scikit-learn to generate a single numerical value indicating the confidence or likelihood that the text to be checked (i.e., response, summary, or single sentence) is hallucinated.   

\subsection{GPT4-based rewriting}
With AI feedback from different sources, we formulated prompts based on a pre-defined template to guide GPT-4 in correcting hallucinations. We explored two distinct rewriting prompts:

\textbf{Rewriting prompt v1.}  This does an exhaustive CoT reasoning or analysis to identify hallucinations in the text. It then does a complete rewriting to correct the hallucination while maintaining the coherence.

\textbf{Rewriting prompt v2.}  This prompt reduces the extent of CoT reasoning and opts to skip rewriting if no hallucinations are detected. When rewriting is necessary, it focuses solely on the hallucinated sentences rather than the entire text. This approach ensures only the essential changes are made to the original content.
  



\section{Main Results}

\subsection{Detection Results}
We compare the different detection methods' performance on the following datasets:

\textbf{Internal benchmark dataset.}
    This internal benchmark dataset has N=1400 examples consisting of 200 representative documents/transcripts x 7 systems of summaries. The ground-truth label is collected by our hired independent data vendors.
    
\textbf{Public benchmark dataset: SummAC.} This dataset, as detailed in \cite{laban-etal-2022-summac} has N=1700 examples collected from six datasets focused on summary inconsistency detection, with ground-truth labels provided by humans in each datasets. We also reference the best results from that paper as baselines.

We evaluated our detection methods against GPT-4 with a focus on both accuracy and latency. As shown in Table \ref{tab:comparing_methods}, NER tends to have lower recall but higher precision in public SummAC \cite{laban-etal-2022-summac}. It's notable that of the three tailored models, the \textbf{SBD} method outperforms the rest in all metrics, showcasing the effectiveness of detection at the token level.
\begin{table}[!ht]
    \centering
\includegraphics[width=0.9\linewidth]{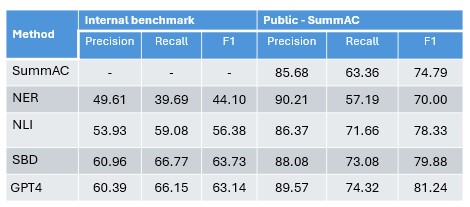}
    \caption{Performance of different methods on the internal benchmark and public SummAC. \textbf{SBD }methods is very competitive and worth further exploring.}
    \label{tab:comparing_methods}
\end{table}
In Table \ref{tab:latency_detection}, Compared with latency of GPT4 on the internal benchmark dataset, our finetuned NLI and SBD models enjoy significant latency advantages.
\begin{table}[!h]
    \centering
    \begin{tabular}{l|l}
        \hline
        Method & Latency (s/request) \\
        \hline
        NLI model& 1.2s (on V100 GPU) \\
        SBD model & 1.3s (on V100 GPU) \\
        GPT4 w/ CoT & 7.9s (per API call)\\
        \hline
    \end{tabular}
    \caption{Latency of finetuned models and the GPT4-based method.}
    \label{tab:latency_detection}
\end{table}
\subsection{Mitigation Results}
Here is the evaluation setting:

\textbf{Benchmark dataset.} This is an internal testset of 200 samples, derived from an application or feature team. This dataset includes 100 documents with two summaries from genearated GPT4 per document: one in paragraph format and one in bullet-point format.

\textbf{Metrics.} We evaluated performance using two key metrics: \textit{mitigation rate} measures the percentage of hallucination being successfully corrected fixed, as determined by GPT4-detection; and \textit{GPT-4 output tokens length} is serving for as proxy metrics for evaluating the latency and cost. 

As shown in Table \ref{tab:mitigation-results}, Rewriting Prompt v2 is great  token efficiency and achieves good a balanced trade-offce among rewriting quality, latency, and cost.

\begin{table}[!h]
    \centering
    \begin{tabular}{l|l|l}
        \hline
        \begin{tabular}{@{}l@{}}Rewriting \\ Prompt \end{tabular}& \begin{tabular}{@{}l@{}}Mitigation \\ Rate \end{tabular}& \begin{tabular}{@{}l@{}}GPT-4 Output \\ Tokens (avg)\end{tabular}   \\
        \hline
        No rewrite & 0.0 & 244 (original output) \\
        Prompt v1 & 66.0\% & 587 \\
        Prompt v2 & 44.7\% & 130\\
        \hline
    \end{tabular}
    \caption{Rewriting to balance the quality and latency.}
    \label{tab:mitigation-results}
\end{table}

\subsection{Performance in Production}
\begin{figure}[h!]
    \centering
    \includegraphics[width=0.95\linewidth]{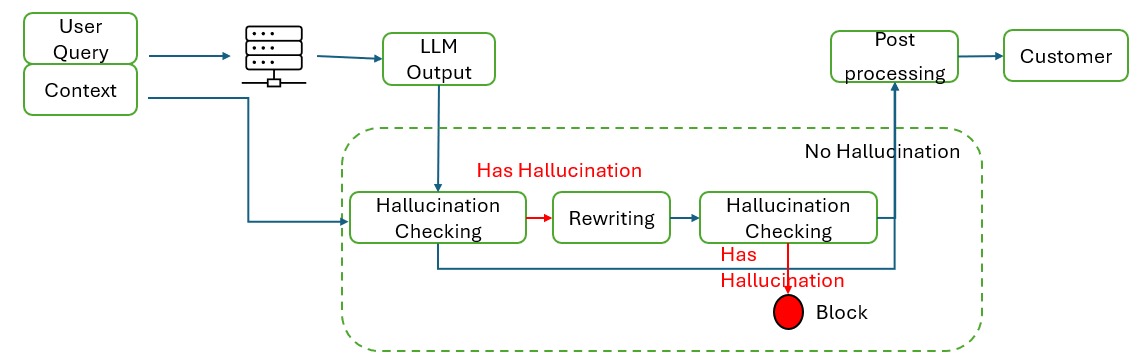}
    \caption{User experience of Hallucination Detection \& Mitigation Pipeline in Production}
    \label{fig:user-experience}
\end{figure}

We have developed two pipelines for production usage: 1) the detection-only pipeline and 2) the detection with mitigation pipeline. They have been integrated into LLM-based products to mitigate the customer's complaints about hallucination as in Figure \ref{fig:user-experience}. In practice, we start with the detection-only pipeline and block the hallucinated context from affecting the customers. Gradually, we move on to the full pipeline of detection and mitigation. 

Accurately measuring the effectiveness of hallucination detection and mitigation in real-world applications is a complex task. To address this, we adopt two approaches: 1) offline human evaluation using a production-related benchmark dataset and 2) GPT4-based online monitoring based on the actual production traffic. 

For the offline human evaluation, we applied our detection and rewriting pipeline to a production-related testset consisting of bullet-point style summaries comprising of 630 individual keypoints. We have 25 (i.e. 4.0\% of 630) in total detected as containing hallucinations. An independent human data labeler to verified that 15 of the 25 dectections were accurate (i.e. check precision is \textbf{60.0\%}). Additionally, the human labeler to check if rewriting fixes the hallucinations, with 10 out 15 are accurately fixed (i.e. rewriting effective rate is \textbf{66.7\%})

In the online monitoring approach, we sampled a portion of production traffic - comprising pairs of <LLM input, LLM ouput > - for evaluation. GPT4 checked outputs for hallucinations and assessed if the rewriting process was necessary. Using GPT4's judgement as a benchmark, we can observed that the precision of ensemble detection is \textbf{above 80\%} (i.e. 80\% of the time is consistent with GPT4's judgement) when the detection rate is about 3\% and rewriting success rate is \textbf{above 50\%}. In Table \ref{tab:online-mitigation-result}, we use statistics based on one-month production traffic to show both the detection-only pipeline and full pipeline of detection and mitigation can effectively reduce hallucinations, albeit an  acceptable increase in latency.

\begin{table}[h!]
    \centering
    \includegraphics[width=0.95\linewidth]{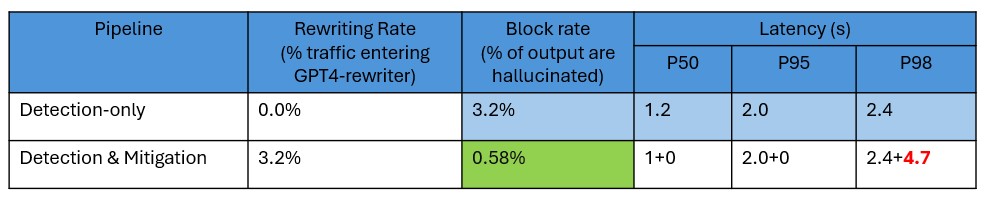}
    \caption{Online monitoring and comparisons of the pipelines based on production traffic.}
    \label{tab:online-mitigation-result}
\end{table}

\section{Challenges and Future Work }



\subsection{Measurement of Effectiveness in Production}
Accurately measuring the effectiveness of hallucination detection and mitigation, as well as the value they bring to customers in a production environment, is very challenging.  We have designed a system that applies mirrored traffic to various pipelines and uses GPT4 to assess hallucination rates and the overall quality of rewritten content. However, the GPT4-based measurement has limitations and ensuring the reliability of these measurements and their alignment with human judgment remains an ongoing challenge, necessitating continuous refinement and validation.

\subsection{Handling Multilingual and Long Source Documents} 
We have incorporated major non-English training datasets into our NLI and SBD models to support multilingual use cases and are utilizing a segmentation-based approach to manage long source documents. However, handling inputs and outputs in different languages and their extensive combinations remains challenging. Additionally, developing effective models for processing long source documents continues to be an open research problem, requiring further exploration and innovation.

\subsection{Deep Customization Needs for Hallucination Handling} 

Different user circumstances call for tailored adjustments to hallucination handling. For example, to meet the production needs, we've calibrated our ensemble-based detection for greater precision with a reduced block rate (or trigger rate) to avoid the availability issue, while also adjusting our rewriting for decreased latency at the cost of some mitigation power. However, there might be another setting, where we need a different balance of quality, latency and cost. Also, adapting to special domain or task or handling a specific types of hallucinations can also be great directions to explore.

\section{Conclusions}


In this paper, we introduce a novel framework that can detect and mitigate intrinsic hallucinations, characterized by outputs not supported by grounding documents in LLMs. Our detection approach leverages the combined strengths of NER, NLI, and novel sequence labelling (SBD), and Decision Tree to detect as much as hallucination as possible. We further developed an effective LLM-based mitigation solution that blance the quality and latency. 




We detail the core elements of our framework and underscore the paramount challenges tied to response time, availability, and performance metrics, which are crucial for real-world deployment of these technologies. Our extensive evaluation, utilizing offline data and live production traffic, confirms the efficacy of our proposed framework and service.

\section{Ethical Considerations} 

Ethical considerations are paramount in the development and deployment of hallucination mitigation systems. 
Ensuring transparency in detection and mitigation processes, providing clear explanations for decisions, and safeguarding user data are essential components of our ethical framework. Balancing these ethical imperatives with technical and operational demands is a complex but necessary challenge. 


\bibliography{anthology,custom}

\appendix



\end{document}